\pdfoutput=1

\documentclass[11pt]{article}

\usepackage{acl}

\usepackage{times}
\usepackage{latexsym}

\usepackage[T1]{fontenc}

\usepackage[utf8]{inputenc}

\usepackage{microtype}

\usepackage{graphicx}
\usepackage{CJKutf8}
\usepackage{stfloats}
\usepackage{multirow}
\usepackage{bm}
\usepackage{amsmath}
\usepackage{array}
\usepackage{booktabs}
\usepackage{makecell}

\newcommand{\chinese}[1]{\begin{CJK}{UTF8}{gkai}{}#1\end{CJK}}
\newcommand{\PreserveBackslash}[1]{\let\temp=\\#1\let\\=\temp}
\newcolumntype{C}[1]{>{\PreserveBackslash\centering}p{#1}}
\newcolumntype{R}[1]{>{\PreserveBackslash\raggedleft}p{#1}}
\newcolumntype{L}[1]{>{\PreserveBackslash\raggedright}p{#1}}

\title{YACLC: A Chinese Learner Corpus with Multidimensional Annotation}

\author{Yingying Wang\textmd{\textsuperscript{1}},
	Cunliang Kong\textmd{\textsuperscript{1}},
	Liner Yang\textmd{\textsuperscript{1}\Thanks{~Corresponding author: L. Yang (\href{mailto:lineryang@gmail.com}{lineryang@gmail.com})}}~,
	Yijun Wang\textmd{\textsuperscript{1}},
	Xiaorong Lu\textmd{\textsuperscript{1}},\\
	\textbf{Renfen Hu\textmd{\textsuperscript{3}},
		Shan He\textmd{\textsuperscript{4}},
		Zhenghao Liu\textmd{\textsuperscript{5}},
		Yun Chen\textmd{\textsuperscript{6}},
		Erhong Yang\textmd{\textsuperscript{1}},
		Maosong Sun\textmd{\textsuperscript{2}}
	} \\
	\textsuperscript{1}School of Information Science, Beijing Language and Culture University \\
	\textsuperscript{2}Department of Computer Science and Technology, Tsinghua University \\
	\textsuperscript{3}Institute of Chinese Information Processing, Beijing Normal University \\
	\textsuperscript{4}Yunnan Chinese Language and Culture College, Yunnan Normal University \\
	\textsuperscript{5}Department of Computer Science and Technology, Northeastern University \\
	\textsuperscript{6}School of Information Management \& Engineering, Shanghai University of Finance and Economics}

\begin{document}
\maketitle

\begin{abstract}
Learner corpus collects language data produced by L2 learners, that is second or foreign-language learners.
This resource is of great relevance for second language acquisition research, foreign-language teaching and automatic grammatical error correction.
However, there is little focus on learner corpus for Chinese as Foreign Language (CFL) learners.
Therefore, we propose to construct a large scale, multidimensional annotated Chinese learner corpus.
To construct the corpus, we first obtain a large number of topic-rich texts generated by CFL learners.
Then we design an annotation scheme including a sentence acceptability score as well as grammatical error and fluency-based corrections.
We build a crowdsourcing platform to perform the annotation effectively\footnote{
\url{https://yaclc.wenmind.net}}.
We name the corpus \textbf{YACLC} (\textbf{Y}et \textbf{A}nother \textbf{C}hinese \textbf{L}earner \textbf{C}orpus) and release it as part of the CUGE benchmark\footnote{\url{http://cuge.baai.ac.cn}}.
By analyzing the original sentences and annotations in the corpus, we found that YACLC has a considerable size and very high annotation quality.
We hope this corpus can further enhance the studies on Chinese International Education and Chinese automatic grammatical error correction. 
\end{abstract}

\section{Introduction}
Learner language is what learners say or write when they are trying to communicate spontaneously in a language they are learning \cite{Corder1967The, Lightbown2013How}.
Currently, there are plenty of studies on error annotation of English learners, and several corpora have been released\cite{Nicholls1999TheCL, Napoles2019Enabling}.
These corpora have not only helped studies on second language acquisition (SLA) , but also applied to grammatical error correction (GEC) tasks \cite{htut-tetreault-2019-unbearable}.

However, there is only few work concentrating on Chinese learner languages, and there is no large-scale Chinese learner corpus providing error corrections.
The widely used HSK Dynamic Composition Corpus \cite{Zhang2009HSK, Zhang2010hanyu} has a problem of unevenly data distribution.
The corpus has a large proportion of data composed of Asian students' essays.
In contrast, the number of essays in European and American countries is relatively small.
The Global Chinese Interlanguage Corpus constructed by \citet{Zhang2013quanqiu} only annotated errors in Chinese characters, words and phrases.
\citet{Wang2015The} constructed the Jinan Chinese Learner Corpus, which annotated the meta-info of learners, such as the author ID, gender, age, education level, L1 language, etc, but didn't annotate the errors.

Therefore, in this work, we propose a large-scale, high-quality Chinese learner corpus, \textbf{YACLC} (\textbf{Y}et \textbf{A}nother \textbf{C}hinese \textbf{L}earner \textbf{C}orpus).
We collect and annotate 32,124 sentences written by CFL learners from the lang-8 platform.
Each sentence is annotated by 10 annotators.
After post processing, a total of 469,000 revised sentences are obtained.

To provide guidance to annotators, we design a multi-dimensional annotation scheme.
As \citet{Zhang2013guanyu} pointed, the same grammatical error may have different properties and types from different perspectives, and hence may have different revisions.
Our scheme asks the annotators to revise from two perspectives of grammar and fluency respectively.

Specifically, following the annotation scheme we designed, the annotator first evaluates the acceptability of a sentence, i.e., whether it is grammatically correct.
If there exists grammatical errors, the annotator need to revise them following the \textit{minimal edits} principle.
Otherwise, the annotator focuses on making sentences more fluent and native-sounding.
The former is called grammatical error correction, and the latter is fluency-based correction.
In YACLC, each original sentence correspond to a binary acceptability score and several grammatical corrections as well as fluency corrections.

We hope this corpus can further enhance the researches on Chinese International Education and Chinese automatic grammatical error correction \cite{ren-2018-sequence, chencheng-2020-chinese}.
It can benefit Chinese teachers to analyze the errors generated by CFL learners, and benefit further teaching.
It also bridges the semantics that learners want to convey with the surface form of actual expressions, providing a chance for automatic grammatical error correction models to better understand the learners' intent and improve the fluency of the final revised sentence.

Therefore, the contributions of this work lie in:

\begin{itemize}
	\item Release a large-scale, high quality Chinese learner corpus with abundant annotations;
	\item Perform multidimensional crowdsourcing annotation on the released corpus.
\end{itemize}

\section{Related Work}
There have been many English learner corpora used in studies of SLA and GEC \cite{liu-2021-neural, yang-2022-controllable}.
The Cambridge Learner Corpus \cite{Nicholls1999TheCL} helped teachers, dictionary editors, English teaching and research experts to study the SLA rules of English learners.
Besides, it also benefits GEC tasks by providing training and testing datasets \cite{htut-tetreault-2019-unbearable}.
\citet{Napoles2019Enabling} proposed \textit{JFLEG}, a parallel corpus which aims to provide fluency edits to make the original text more native sounding.
On the other hand, the construction of the Chinese learner corpus is still in development.
\citet{Zhang2009HSK, Zhang2010hanyu} collected and annotated the composition answer sheets in high-level Chinese Proficiency Test (HSK), with a total of 4.24 million words.
The Global Chinese Interlanguage Corpus \cite{Zhang2013quanqiu} has a larger scale and a richer source, and currently has 107 million words.
\citet{zhao-2018-nlpcc} collected a Chinese Mandarin learner corpus by exploring "language exchange" social networking services (SNS).
By collecting essays written by CFL learners and revised version by Chinese natives, the corpus contains 717,241 sentences from writers.
The corpus is then used to hold a shared task of Chinese GEC in NLPCC 2018.

In this work, we also collect sentences written by learners on SNS websites.
Differently, we design a multidimensional annotation scheme and perform crowdsourcing annotation on a self-build platform.
Considering the two principles of \textit{minimal edits} \cite{ng-etal-2014-conll} and \textit{fluency edits} \cite{Napoles2019Enabling}, our scheme ask annotators to write multiple \textit{correct} sentences from different perspectives.
And each revision has a tag of grammatical or fluency, which identifies the different principles it follows.

\section{Corpus Construction}
We construct the YACLC in three steps, i.e., data collection, annotation, and post-processing.
This section introduces these steps one by one.

\subsection{Data Collection}
We collect sentences written by CFL learners from \url{http://lang-8.com}, a language-learning website where learners freely sharing their essays.
We follow \citet{mizumoto-etal-2011-mining} to explore the "language exchange" social networking service.
The collected data contains 441,670 sentences from 29,595 essays, provided by around 50,000 CFL learners.

As the essays on the lang-8 are not unified and there is lots of noise in raw sentences, we take a series of measures to clean up the data.
Firstly, based on user information listed on the website, we filter out essays written by Chinese native speakers.
As for essays written in traditional Chinese, we use the OpenCC\footnote{\url{https://github.com/BYVoid/OpenCC}} toolkit to change them into simplified Chinese.
In order to make the data more compact, we remove repetitive sentences and rather simple sentences, such as \chinese{大家好}(\textit{Hello everyone}) and \chinese{晚上好}(\textit{Good evening}).
In addition, learners may forward good Chinese articles or lyrics.
In such cases, some keywords may appear in the text, such as \chinese{转载}(\textit{Forward}).
We eliminate texts that contains these specific keywords.

After the clean up, we obtain 32,124 sentences from 2,421 essays. 
We upload these sentences to the crowd-sourcing platform, and then organize annotators to revise them.

\begin{figure*}[t]
	\centering
	\includegraphics[width=0.9\linewidth]{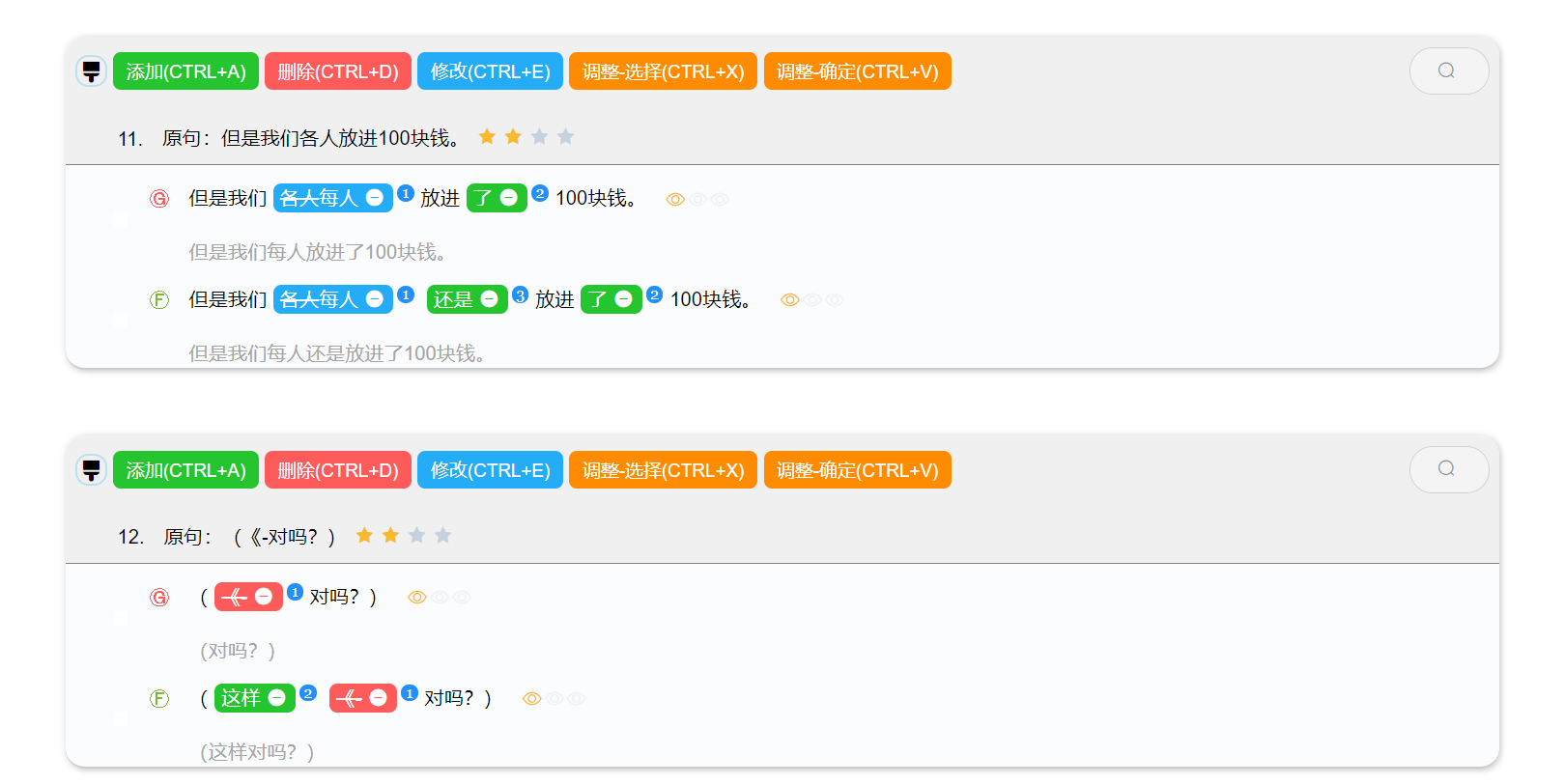}
	\caption{The crowdsourcing Chinese learner text annotation platform.}
	\label{fig:platform}
\end{figure*}

\subsection{Annotation}
We recruit 183 annotators to revise the collected sentences.
To adopt the crowd-sourcing strategy, we build a Chinese learner text annotation platform, as shown in Figure \ref{fig:platform}.

We first introduces our annotation principles, then our designed scheme as well as the platform's functions.

\subsubsection{Principle}
Our goal in this work is to construct a large-scale, high-quality Chinese learner corpus, providing both grammatical revisions and fluent revisions.
Referring to previous work of English Learner Corpus \cite{ng-etal-2013-conll, ng-etal-2014-conll, Napoles2019Enabling} and considering the unique features in Chinese, we formulate the following principles.
These principles are carried out in the entire annotation process.

\begin{itemize}
	\item
	\textbf{The minimal annotation unit is word.}
	Unlike languages such as English, there is no natural separation between Chinese words.
	But since words are the smallest language unit that can be used independently, the minimal unit of revision should be word rather than character.
	For example, if the word \chinese{足球}(\textit{football}) is written as \chinese{足求}(\textit{foot-beg}), the entire word should be annotated rather than just the character \chinese{求}(\textit{beg}).
	
	\item
	\textbf{Be faithful to the original intention.}
	The annotation should be faithful to the intention of CFL learners.
	The original semantics should be retained as much as possible \cite{Zhang2013guanyu}.
	
	\item
	\textbf{Multi-dimensional annotation.}
	For each original sentence, an annotator may give a variety of revisions.
	Each revision should be a grammatical or fluency correction, and marked which one it belongs to.
	The grammatical correction is to make the sentence conform to grammar, following the \textit{minimal edits} \cite{ng-etal-2014-conll} principle.
	The fluency correction is to make the sentence \textit{fluent} and native-sounding.
\end{itemize}

\subsubsection{Scheme}
The annotation process is divided into two stages, which are acceptability scoring and sentence revision.

\paragraph{Acceptability Scoring}
The acceptability score indicates whether the sentence has grammatical errors.
If there is, the score is 0, and the annotator needs to make at least one grammatical correction.
If not, the score is 1, and the annotator can only make fluency corrections.

\paragraph{Sentence Revision}
Following \citet{Lu1994waiguo}, we classify the errors in original sentences into four types, which are missing components, redundant components, misuse of word, and wrong word order.
Corresponding to these types, we design four operations to modify them, namely add, delete, replace, and adjust order.
Both grammatical and fluency corrections are implemented through these four basic operations.

\subsubsection{Crowdsourcing Platform}
As shown in Figure \ref{fig:platform}, we implement the platform for crowdsourcing annotation.
The interface of the platform is friendly designed and very easy to use.
The four stars following the original sentence indicate acceptability, and more than two stars are regarded as 1, otherwise it is regarded as 0.
We design four operation buttons according to the scheme.
To make a revision, an annotator need to choose the word to be revised, and then click the corresponding button.
Our platform also has four hot-keys for efficient operation.
Rather than click the buttons, annotators can press hot-keys instead to speed up annotation.
Each correction has a mark of \textbf{G} or \textbf{F} indicating grammatical or fluency.
Annotators can alter between these two states. 

\begin{table*}[t]
	\centering
	\begin{tabular}{lR{1.5cm}R{1.5cm}R{1.5cm}R{1.5cm}R{1.5cm}R{1.5cm}R{1.5cm}}
		\toprule
		& Sent & Word Token & Char. Token & Word Type & Words per Sent & Chars per Sent & Type/Token Ratio \\
		\midrule
		Original & 32,124 & 373,749 & 577,679 & 23,374 & 11.63 & 17.98 & 15.99 \\
		Grammatical & 331,292 & 3,931,509 & 6,077,742 & 33,042 & 11.87 & 18.35 & 118.99 \\
		Fluency & 137,708 & 1,405,090 & 2,172,490 & 24,240 & 10.20 & 15.78 & 57.97 \\
		\bottomrule	
	\end{tabular}
	\caption{Statistics of the annotations.}
	\label{table:statistics}
\end{table*}

\begin{figure}[t]
	\centering
	\includegraphics[width=\linewidth]{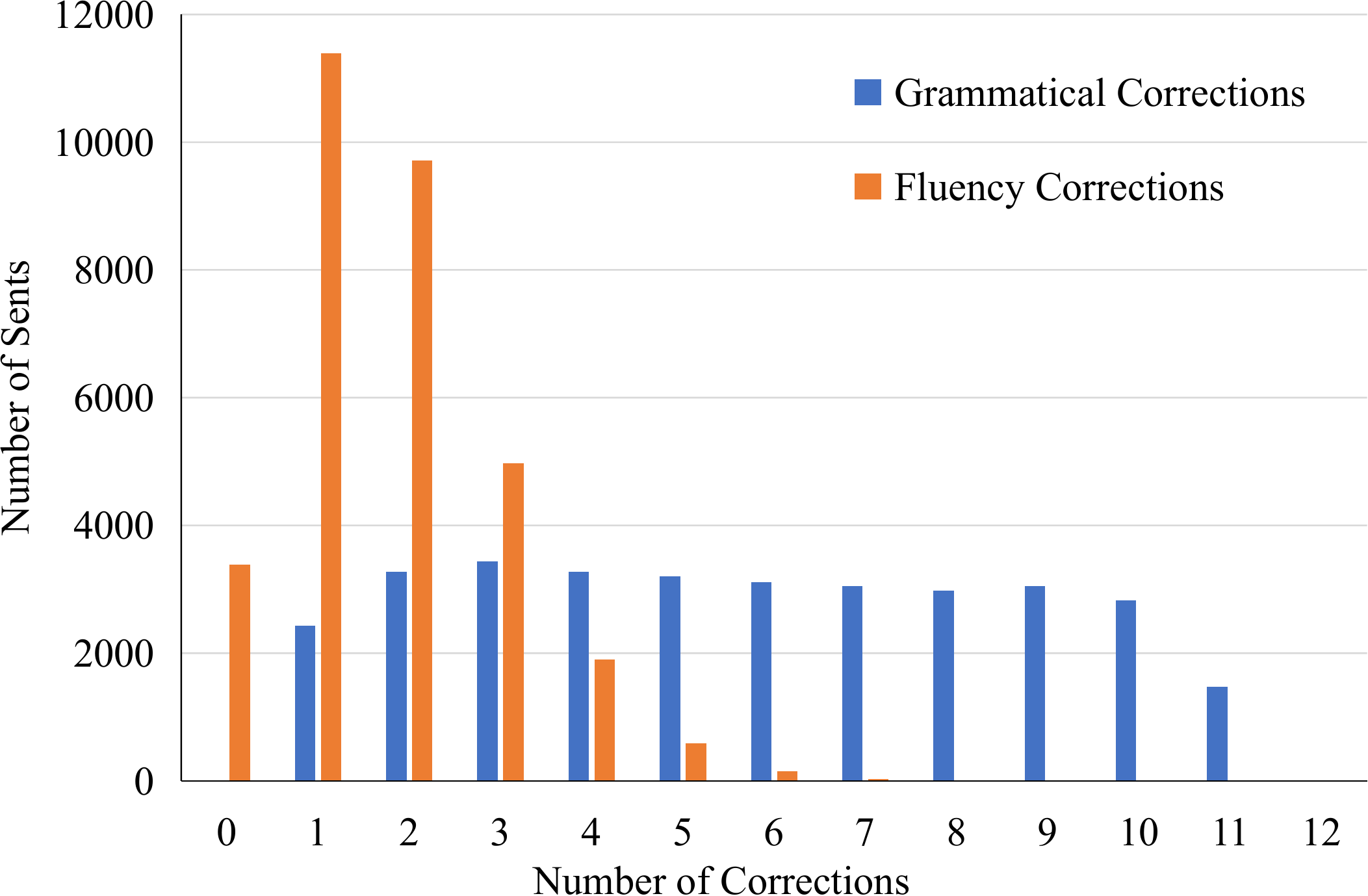}
	\caption{Distribution of corrections. the blue bars are grammatical corrections, and the orange bars are fluency corrections.}
	\label{fig:dist}
\end{figure}

\subsection{Post-Processing}
After the annotation, we collect the acceptability scores and revisions on the platform, and then deduplicate the same revisions.
Finally, we organize the corpus into the \textit{json} format.

\section{Corpus Analysis}
This section shows the statistics of YACLC and evaluates its quality.

\subsection{Basic Statistics}
The annotated YACLC is a large-scale corpus.
As shown in Table \ref{table:statistics}, we annotated 469,000 revisions for 32,124 sentences written by CFL learners, 14.6 revisions per sentence on average.

\subsection{Correction Analysis}
We analyze the distribution of both grammatical and fluency corrections.
As shown in Figure \ref{fig:dist}, the grammatical corrections has a more balanced distribution.
However, the distribution of fluency corrections skews to the left.
This indicates that comparing with grammatical corrections, annotators write fewer fluency corrections in general.
This is also in line with our intuition, because there can be many ways to correct grammatical errors, but only one native-sounding fluent expression.

\begin{figure}[t]
	\centering
	\includegraphics[width=\linewidth]{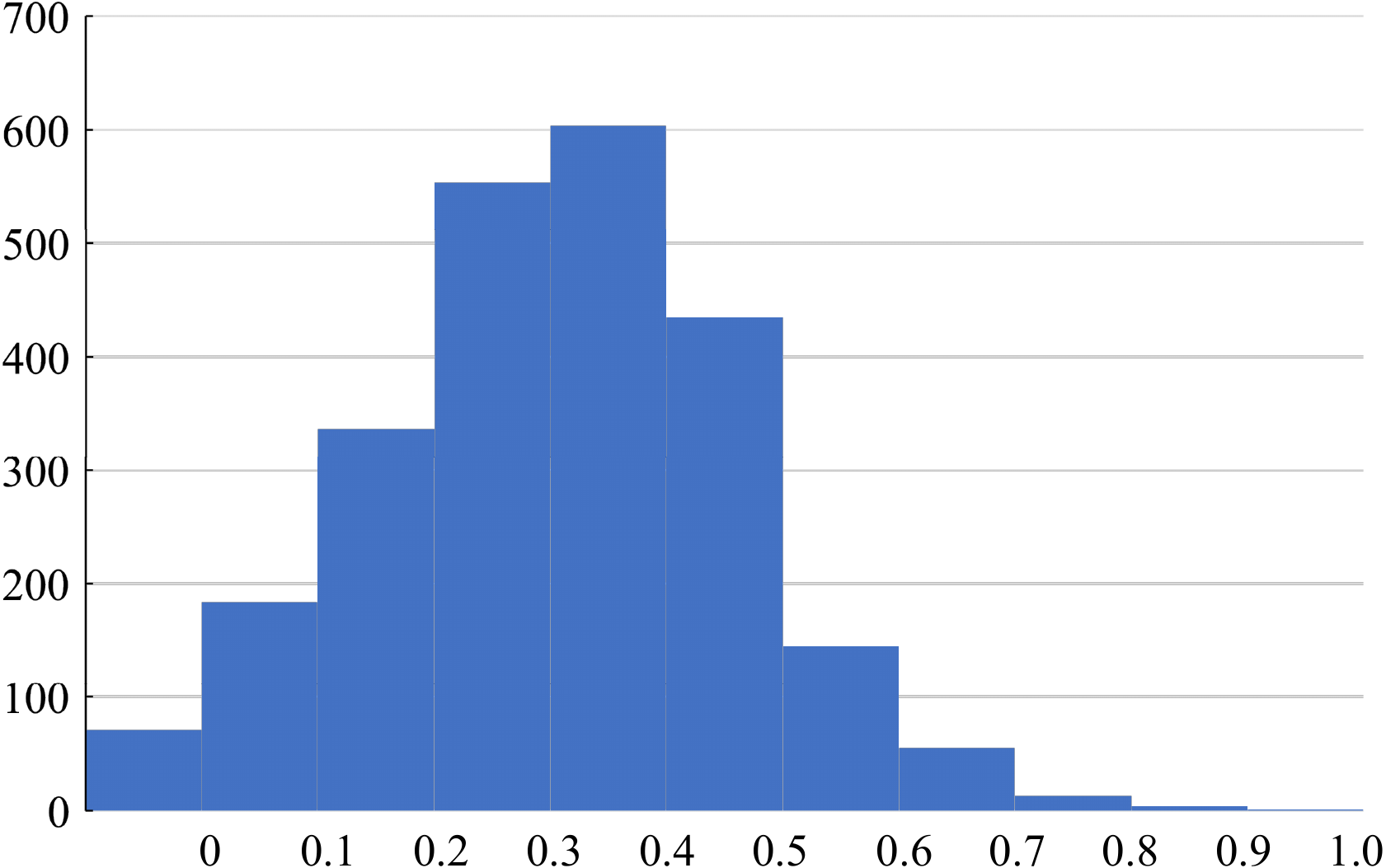}
	\caption{Cohen's kappa coefficient indicating consistency of the acceptability score.}
	\label{fig:consist}
\end{figure}

\subsection{Acceptability Analysis}
We illustrate the consistency of acceptability score in Figure \ref{fig:consist}.
The average Kappa score is 0.38, which indicates a fair consistency.

\section{Conclusion}
In this work, we construct a large-scale, high-quality Chinese learner corpus, YACLC.
We perform multidimensional crowd-sourcing annotation to build this corpus.
After annotation, the corpus contains both grammatical and fluency corrections for sentences written by CFL learners.
We hope that YACLC can further enhance the researches on Chinese International Education and the task of Chinese automatic grammatical error correction.

\bibliography{ccl2020}
\bibliographystyle{acl_natbib}

\end{document}